# PHYSICS INFORMED NEURAL NETWORK FOR CONCRETE MANUFACTURING PROCESS OPTIMIZATION

Sam Varghese
MBA, BTech (CE), NMIMS
Intern, Industry 4.0, KPMG India
sam.varghese08@nmims.in

Rahul Anand
Associate Director, KPMG India
rahulanand2@kpmg.com

Dr. Gaurav Paliwal
Assistant Professor, NMIMS
gaurav.paliwal@nmims.edu

## ABSTRACT

Concrete manufacturing projects are one of the most common ones for consulting agencies. Because of the highly non-linear dependency of input materials like ash, water, cement, superplastic, etc; with the resultant strength of concrete, it gets difficult for machine learning models to successfully capture this relation and perform cost optimizations. This paper highlights how PINNs (Physics Informed Neural Networks) can be useful in the given situation. This state-of-the-art model shall also get compared with traditional models like Linear Regression, Random Forest, Gradient Boosting, and Deep Neural Network. Results of the research highlights how well PINNs performed even with reduced dataset, thus resolving one of the biggest issues of limited data availability for ML models. On an average, PINN got the loss value reduced by 26.3% even with 40% lesser data compared to the Deep Neural Network. In addition to predicting strength of the concrete given the quantity of raw materials, the paper also highlights the use of heuristic optimization method like Particle Swarm Optimization (PSO) in predicting quantity of raw materials required to manufacture concrete of given strength with least cost.

## 1. INTRODUTION

Manufacturing concrete is a complex procedure with a lot of chemical reactions and physics laws determining the strength of resultant concrete. [1] The raw materials generally required for its manufacturing are cement, slag, ash, water, superplastic, coarse aggregate and fine aggregate. The strength of concrete thus produces is a highly non-linear function of the quantity of raw materials passed. This makes it extremely difficult to get the models perform accurate predictions and thus help with cost optimization. The objective of the paper shall be:

1. To create advance machine learning models capable of capturing the highly non-linear dependency of concrete strength with quantity of raw materials used, thus making us capable of predicting the resultant strength.
2. To integrate above mentioned model with heuristic optimization algorithm for calculating quantity of raw materials required to produce concrete of least input strength.

## 1.1 Background Knowledge

In concrete manufacturing, [2] Portland cement is the primary binding agent used. The production process of Portland cement begins with the extraction and crushing of limestone, which is then blended with clay and other aggregates. Primary compounds of Portland cement are:

- Tricalcium Silicate ($3CaO.SiO_2$ or $C_3S$)
- Dicalcium Silicate ($2CaO.SiO_2$ or $C_2S$)
- Tricalcium Aluminate ($3CaO.Al_2O_3$ or $C_3A$)
- Tetracalcium Aluminoferrite ($4CaO.Al_2O_3.Fe_2O_3$ or $C_4AF$)

In these formulas, C stands for calcium oxide (lime), S for silica, A for alumina, and F for iron oxide.

In a cement kiln, this mixture is ground into fine powder in a high temperature, where the raw materials undergo a chemical transformation. The kiln operates at temperatures exceeding 1400°C (2552°F), causing the limestone and clay to react and form clinker, a granular material that is the key intermediate product in cement production. The calcination of limestone happens according to the chemical equation (1)

$$CaCO_3 \xrightarrow{yields} CaO + CO_2 \qquad (1)$$

The produced clinker is cooled and ground to a very fine powder. The powdered clinker shall be mixed only with a little primary ingredient of gypsum and sometimes with another material that controls the setting time for the cement. By this grinding process, it then forms a final product called Portland cement.

The cement undergoes a chemical reaction called cement hydration when mixed with water, which leads to the formation of calcium silicate hydrate and other compounds that give the cement its binding properties. The primary reactions that occur at this stage is mentioned in the reaction (2), (3), (4), (5) and (6) given below:

- Tricalcium Silicate

$$2(3CaO.SiO_2) + 6H_2O \xrightarrow{yields} 3CaO.2SiO_2.3H_2O + 3Ca(OH)_2 \qquad (2)$$

- Dicalcium Silicate

$$2(2CaO.SiO_2) + 4H_2O \xrightarrow{yields} 3CaO.2SiO_2.3H_2O + Ca(OH)_2 \qquad (3)$$

- Tricalcium Aluminate

$$C_3A + 6H_2O \xrightarrow{yields} 3CaO.Al_2O_3.6H_2O \qquad (4)$$

- Tricalcium Aluminate with Gypsum

$$C_3A + 3CaSO_4.2H_2O + 26H_2O \xrightarrow{yields} C_6A.S_3H_{32} \qquad (5)$$

- Tetracalcium Aluminoferrite

$$C_4AF + 10H_2O \xrightarrow{yields} 4CaO.Al_2O_3.Fe_2O_3.10H_2O \qquad (6)$$

Portland cement is then combined with aggregates such as sand, gravel, or crushed stone to produce concrete. The mixing process involves combining the cement, aggregates, and water to create a homogeneous mixture. Once mixed, concrete is poured into molds or forms, where it sets and cures to achieve its final strength and durability. The hydration process continues over time, gradually hardening the concrete and allowing it to reach its desired structural properties. This combination of cement, aggregates, and water is crucial in constructing durable and robust structures, making concrete one of the most widely used construction materials in the world.

## 1.2 Our Approach

Both the project objectives will be entirely addressed in sequence. First, machine learning models will be developed to get the strength of the concrete being used. That means to create and evaluate different models, including the Physics-Informed Neural Network as per [3]. For instance, results obtained from the PINN shall be compared using traditional models like Linear Regression, Deep Neural Networks, etc. Feature engineering is another operation carried out to enhance the predictive power in the models. This is a process whereby features are transformed and selected in such a way that they help make the model very effective by reducing the loss value during training. Therefore, the models fit very well and can predict concrete strength adequately.

Once the models are trained and validated for predicting concrete strength, they will be applied to the second objective: optimizing the quantity of raw materials required to achieve a target strength value. This involves using the strength prediction models to estimate the necessary raw materials while considering cost efficiency. The optimization process will employ heuristic methods to identify the best combination of materials that maximizes concrete strength and minimizes procurement costs.

Optimization will be by Particle Swarm Optimization. PSO is a heuristic optimization for finding optimal solutions in complex problem spaces. By applying PSO, the project aims to determine the most cost-effective mix of raw materials that meets the desired strength requirements. This approach ensures that the final concrete mix is both economically viable and meets performance specifications, ultimately balancing strength and cost considerations.

## 1.3 Permutation Feature Importance

[4] In order to find which features have been the most useful for the models that would be trained, permutation feature importance algorithm has been used. Here is how the mentioned algorithm works:

- After training the model M on the dataset D, it's performance P gets noted.
- If the feature whose importance needs to be calculated is F, then the values of F get shuffled randomly. Rest of the features remains untouched.
- Model M gets re-trained on the new dataset, and the performance P` is noted. Because of the random shuffling of F, in almost all the cases P` < P and the difference between both these values would get declared as the importance score of F.

## 2. RELATED WORKS

The following is the work that has been performed in concrete production.

Concrete is one of the highly adaptable and commonly used materials, basically consisting of Portland cement, aggregates, and water. It is in the composition of the Portland cement and the process it will undergo as it hydrates that root most of the distinctiveness of concrete. A proper understanding of the specific compounds within the cement and how they react with water is the leading secret to customizing concrete for different uses and ensuring it performs according to specific criteria.

Interaction of all forms of cement with water initiates a very complex chemical reaction. These chemical reactions are very vital because it leads to the hydration reactions responsible for strength and lifespan of concrete.

The way in which these compounds interact during hydration significantly influences the microstructure of the concrete. Recent advances in modelling these reactions have provided deeper insights into how this process works at a microscopic level. Additionally, the ratio of water to cement is and important factor on which the strength depends.

Recent advancements in predicting concrete strength have leveraged machine learning models, especially artificial neural networks (ANNs). These models can uncover complex relationships between input variables, such as mix proportions and curing conditions, and the resulting concrete strength [15]. Yeh demonstrated the effectiveness of ANNs in enhancing prediction accuracy over traditional methods [16]. Similarly, Chou and Tsai and Ni and Wang explored neural networks and other techniques for strength prediction [17][18], while Gupta and Wu expanded the analysis to include support vector machines and decision trees [19][20].

To optimize the performance of these machine learning models, feature engineering plays a pivotal role. This process involves selecting, transforming, and creating features to improve model accuracy, including techniques such as outlier removal and data normalization [21]. Effective feature engineering is crucial, as demonstrated by comparisons of various methods [22], and contributes to reducing errors in concrete strength predictions [25]. Liu and Motoda and Guyon and Elisseeff emphasize the importance of feature selection for enhancing predictive performance [23][24].

Genetic Algorithms, PSO, etc are instrumental in refining concrete mix designs. These algorithms aim to identify optimal raw material proportions that minimize costs while meeting performance criteria [26]. Kennedy and Eberhart illustrate PSO's success in achieving significant cost reductions and strength improvements [27]. Complementarily, Holland and Goldberg provide insights into genetic algorithms and their application in optimization [28][29]. Various optimization techniques discussed by Deb (1988) further support the development of cost-effective concrete mix designs [30].

Recent advancements, including the use of supplementary cementitious materials (SCMs) like ash and silica, have enhanced concrete properties. These materials perform reactions which helps with improving strength and durability [31]. Hooton discusses the benefits of incorporating ground granulated blast-furnace slag [32], while Thomas and Reinhardt explore fly ash and self-healing concrete technologies [33][34]. The integration of nanomaterials has further propelled advancements in concrete performance [35].

Concrete durability, particularly in aggressive environments, is a critical concern. Carbonation, the reaction of carbon dioxide with calcium hydroxide, affects concrete durability and service life [36]. Papadakis review the characteristics impacting durability with a focus on carbonation [37], and Morandeau and White provide an overview of the carbonation process and its implications [38]. Strategies to enhance durability, such as using pozzolanic materials and proper curing techniques, are also discussed [39][40].

Machine learning applications, including advanced techniques like deep learning, have shown promise in analyzing and predicting concrete properties. These methods offer new ways to handle complex data and enhance prediction accuracy [41]. Foundational texts by Goodfellow on deep learning and Hastie on statistical learning underscore the potential of these techniques in concrete technology [42][43].

Murphy and Alpaydin further explore machine learning applications in engineering, providing powerful tools for advancing concrete research [44][45].

The integration of advanced ML models, such as Physics-Informed Neural Networks (PINNs), with heuristic optimization techniques holds significant promise for advancing concrete technology. By leveraging these sophisticated approaches, it is possible to get better models and optimize the manufacturing processes to reduce costs and enhance the performance of concrete.

## 3. Methodology

### 3.1 Data Availability

The dataset utilized for the research has been acquired from Kaggle [46] and is CC0 licensed, hence all the permissions to utilize it has been granted. Table 1 highlights the structure of the table:

Table 1: Upper 5 rows of the concrete manufacturing dataset

| Cement | Slag | Ash | Water | Superplastic | Coarse Agg. | Fine Agg. | Age | Strength |
|---|---|---|---|---|---|---|---|---|
| 141.3 | 212.0 | 0.0 | 203.5 | 0.0 | 971.8 | 748.5 | 28 | 29.89 |
| 168.9 | 42.2 | 124.3 | 158.3 | 10.8 | 1080.8 | 796.2 | 14 | 23.51 |
| 250.0 | 0.0 | 95.7 | 187.4 | 5.5 | 956.9 | 861.2 | 28 | 29.22 |
| 266.0 | 114.0 | 0.0 | 228.0 | 0.0 | 932.0 | 670.0 | 28 | 45.85 |
| 154.8 | 183.4 | 0.0 | 193.3 | 9.1 | 1040.4 | 696.7 | 28 | 18.29 |

First 7 features of Table 1 are raw materials required to manufacture the concrete. This concrete needs to be set for a few days, ranging from 1 to 28 days. The distribution of age is shown in Figure 1

## 3.2 Research Methodology Flowchart

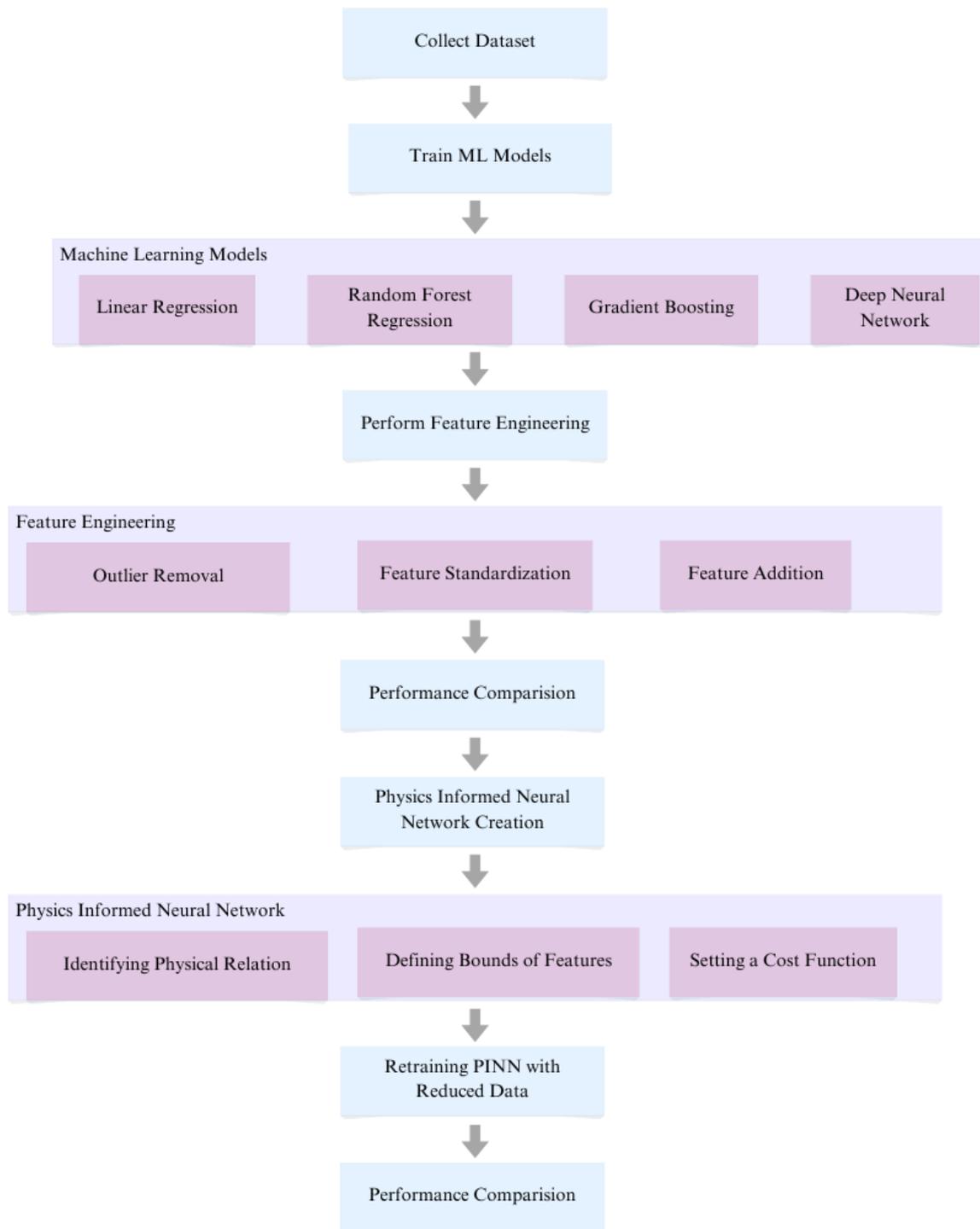

Figure 1: Flowchart of the research

## 4. Implementation

The dataset used in the research has been taken from Kaggle, and is CC0 licensed, hence all the permissions to utilize it has been granted. Table 1 highlights the structure of the table:

Table 1: Upper 5 rows of the concrete manufacturing dataset

| Cement | Slag | Ash | Water | Superplastic | Coarse Agg. | Fine Agg. | Age | Strength |
|---|---|---|---|---|---|---|---|---|
| 141.3 | 212.0 | 0.0 | 203.5 | 0.0 | 971.8 | 748.5 | 28 | 29.89 |
| 168.9 | 42.2 | 124.3 | 158.3 | 10.8 | 1080.8 | 796.2 | 14 | 23.51 |
| 250.0 | 0.0 | 95.7 | 187.4 | 5.5 | 956.9 | 861.2 | 28 | 29.22 |
| 266.0 | 114.0 | 0.0 | 228.0 | 0.0 | 932.0 | 670.0 | 28 | 45.85 |
| 154.8 | 183.4 | 0.0 | 193.3 | 9.1 | 1040.4 | 696.7 | 28 | 18.29 |

First 7 features of Table 1 are raw materials required to manufacture the concrete. This concrete needs to be set for a few days, ranging from 1 to 28 days. The distribution of age is shown in Figure 2

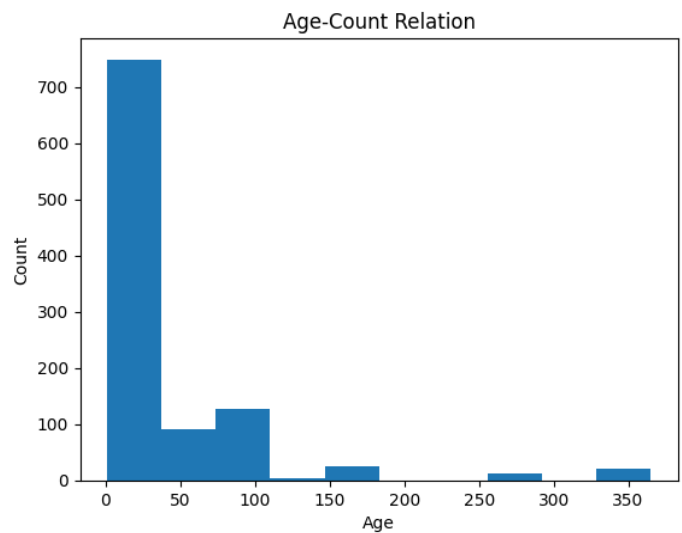

Figure 2: Distribution of concrete age values

From Figure 2, it is evident that age 28 is the most common in the dataset, hence all the analysis is done on the age 28 records. Figure 2 indicates the correlation of features with strength

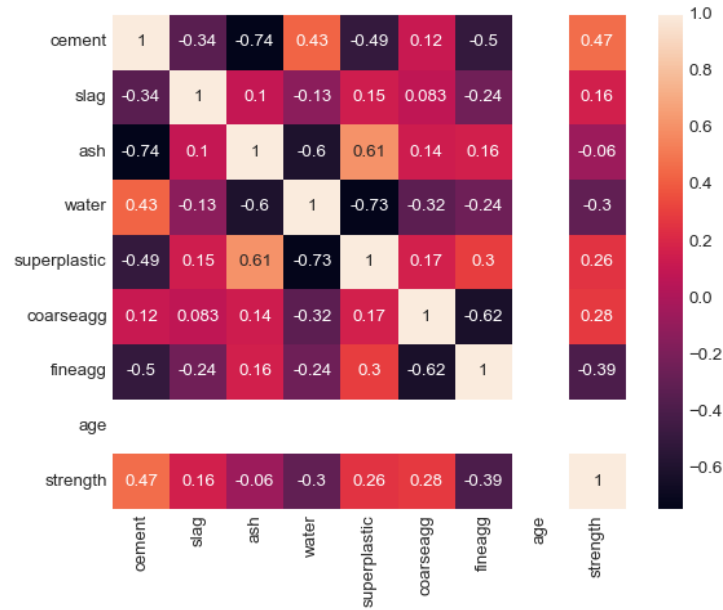

Figure 3: Heatmap of Table 1

From Figure 3, it is evident that cement affects the strength of the resultant concrete the most, followed by fine aggregate. For creating models capable calculating the quantity of raw materials, the focus shall first be on predicting the strength given raw materials quantity. This model will then be used in the heuristic optimization for predicting raw material quantity. The following portion give details of ML models that have been trained on raw dataset. These models have been compared with the ones, trained on feature engineered dataset.

## 5. Strength Prediction Models, Without Feature Engineered Dataset

Preparing models capable of predicting strength, given the quantity of input items is necessary for making particle swarm optimization model work. Initially, all the models mentioned below shall be trained on raw dataset. 90% of the data gets used for training, while the remaining 10% for testing. After feature engineering, the results of models shall get compared. For predicting the strength, the following models have been utilized:

### 5.1 Linear Regression Model

Linear regression model is the most simplistic model of the paper. It fits the equation

$Y = \beta_0 + \beta_1 X_1 + \beta_2 X_2 + \ldots \beta_n X_n + \epsilon$ to the data points. Here, Y represents the dependent variable, X are the independent variables, β are the coefficients that describe the relationship, and ε is the error term. What makes linear regression so widely used is its simplicity and ease of interpretation. It provides clear insights into how changes in the independent variables influence the dependent variable. In this paper, linear regression is the most basic model discussed, and its performance will serve as a benchmark for comparing more complex models that follow.

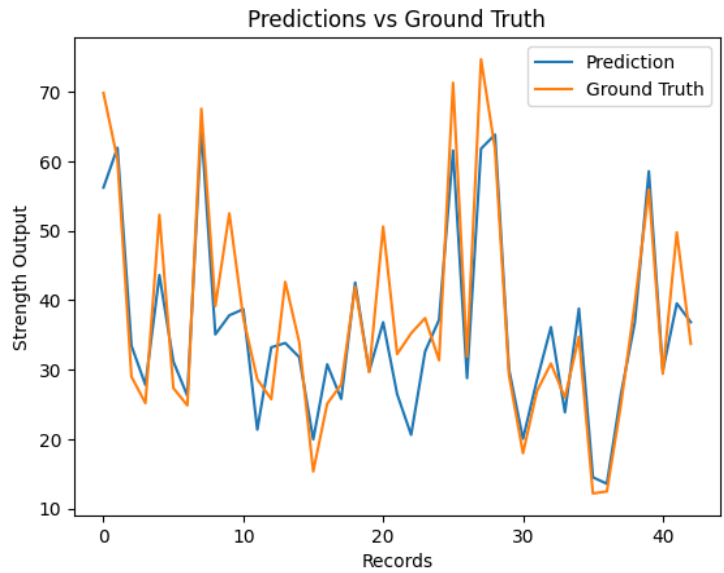

Figure 4: Predicted values vs ground truth

Figure 4 compares the predictions of the linear regression model with the ground truth. For comparing models, RMSE (Root Mean Squared Error) loss value has been used. It's calculated using the formula (7)

$$\text{RMSE} = \sqrt{\frac{\sum_{i=0}^{N}(y_i - \hat{y}_i)^2}{N}} \tag{7}$$

$y_i$ is the ground truth value, while $\hat{y}_i$ is the value predicted by the model. Figure54 indicates the permutation feature importance score of all the features:

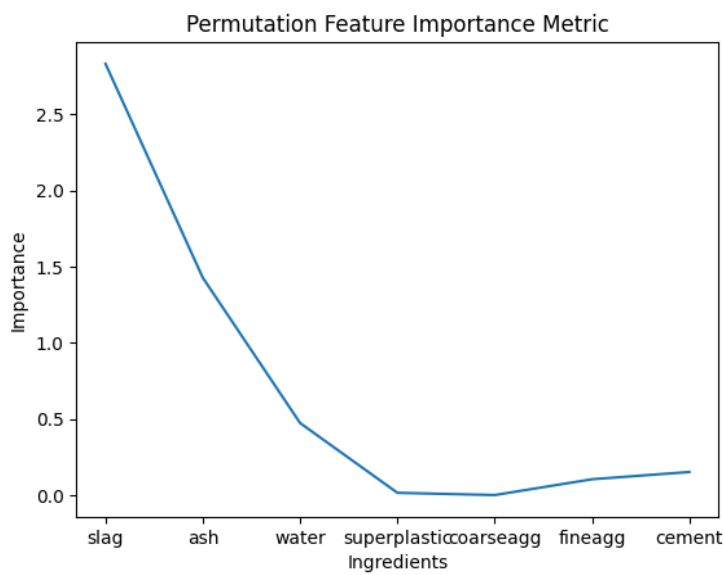

Figure 5: Permutation feature importance of linear regression model

According to the results of feature importance from Figure 5, slag is the input that has the highest importance in determining the strength of concrete, followed by ash, and water.

## 5.2 Random Forest Regression

[47] Random Forest model combines multiple decision trees to perform regression. With this approach, the performance generally tends to increase. When this model gets trained on the raw dataset, unfortunately it fails to perform any better than the simple linear regression model. Figure 6 illustrates the predicted value from this model, and the corresponding ground truth values

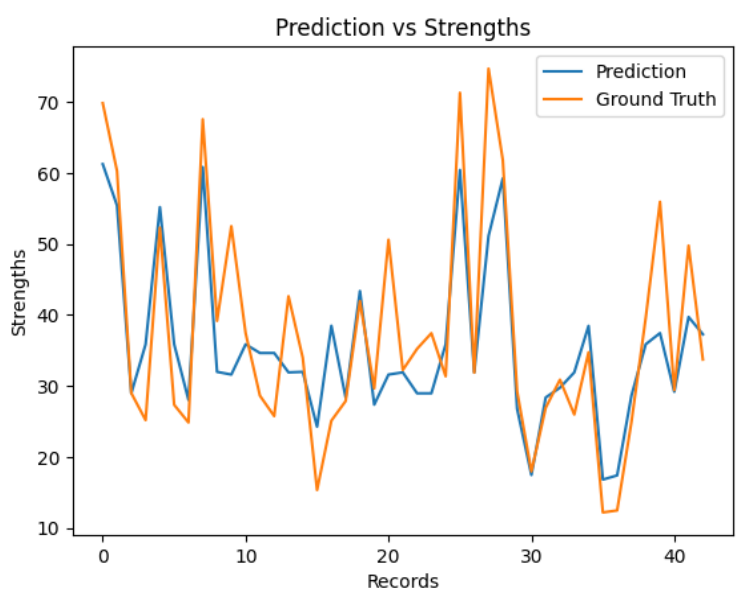

Figure 6: Predicted values and the ground truth values for Random Forest model

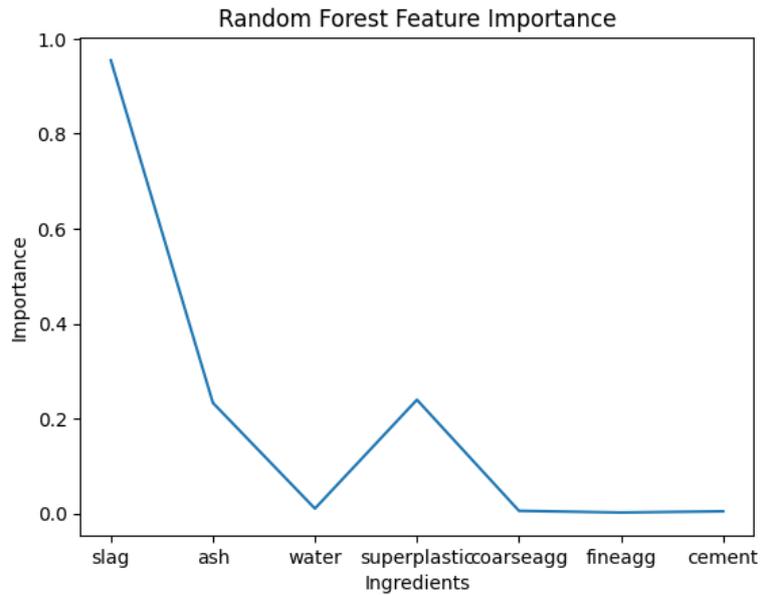

Figure 7: Feature importance, random forest model

### 5.2.1 Hyperparameter Tuning

For the model, the count of trees (*n_estimators)* parameter is set at 10, the maximum depth (*max_depth*) is set at 3, and the random state (*random_state*) is 0. For increasing the model complexity, the value of *n_estimators*, and *random_state* can be increased.

## 5.3 Gradient Boosting

Gradient boosting is another powerful machine learning technique for regression and classification. [48] This technique creates an ensemble of decision trees sequentially in which every new tree corrects the errors of the previous ones. It combines the predictions from multiple weak learners, leading to an improved accuracy of the model. The method minimizes a specified loss function using gradient descent to optimize performance. When trained on the raw dataset, this model gave an RMSE loss value less than random forest, but is still not up to the mark.

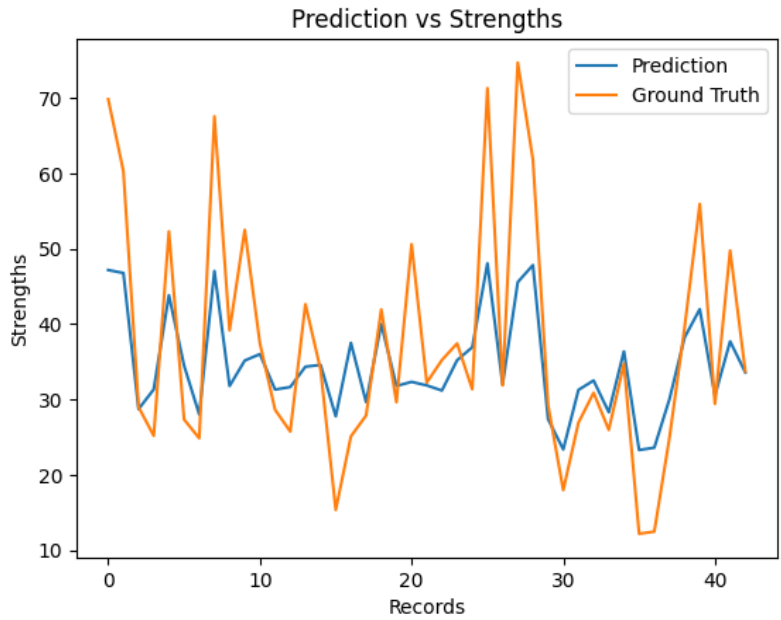

Figure 8: Prediction and corresponding ground truth for Gradient Boosting

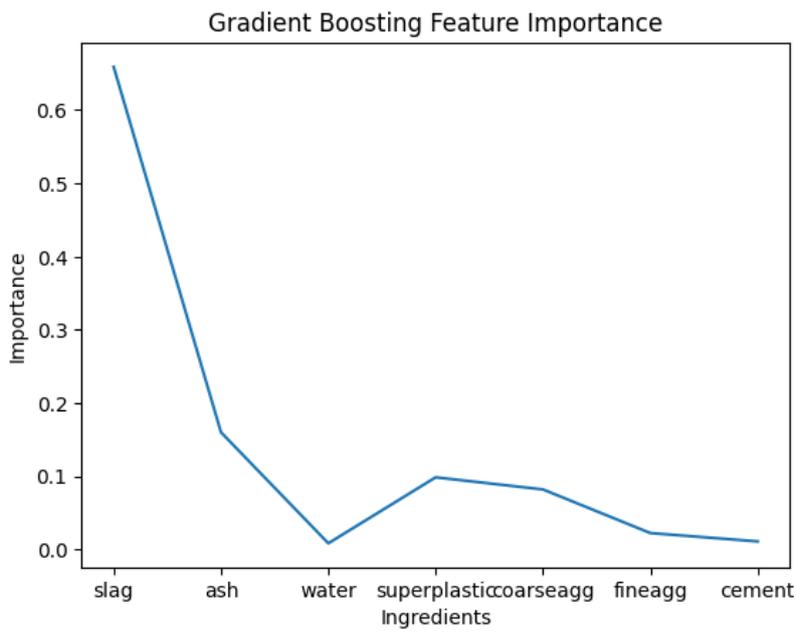

Figure 9: Permutation feature importance from gradient boosting model

### 5.3.1 Hyperparameter Tuning

For Gradient Boosting, Table 2 indicates the values of the hyperparameters used:

Table 2: Random Forest Hyperparameters

| Parameter | Value |
| --- | --- |
| n_estimators | 100 |
| learning_rate | 0.01 |
| max_depth | 5 |
| random_state | 0 |
| min_sample_split | 5 |

## 5.4 Deep Neural Network

[49] Deep Neural Networks are a class of learning models that represent several layers of interconnected neurons that have a way of learning deep patterns and deep representations based on big datasets. Concretely, this means that each layer contains a DNN, which, from the input received at that layer, extracts feature at progressively deeper levels of abstraction, allowing the network to carry out complex tasks from image recognition to natural language processing and playing games. Using massive amounts of data and computational power are the two key factors allowing DNNs to achieve high accuracy and generalization in diverse applications, which makes them one of the vital blocks in modern artificial intelligence and deep learning research.

Because of the unique ability of Deep Neural Networks to capture highly non-linear dependency of features, it's beneficial to give it a try for predicting concrete strength. After training the RMSE loss value was 7.03 which is the least encountered yet. This confirms that the very quality of Neural Networks of capturing the non-linear dependency proved out to be crucial for enhancing the performance.

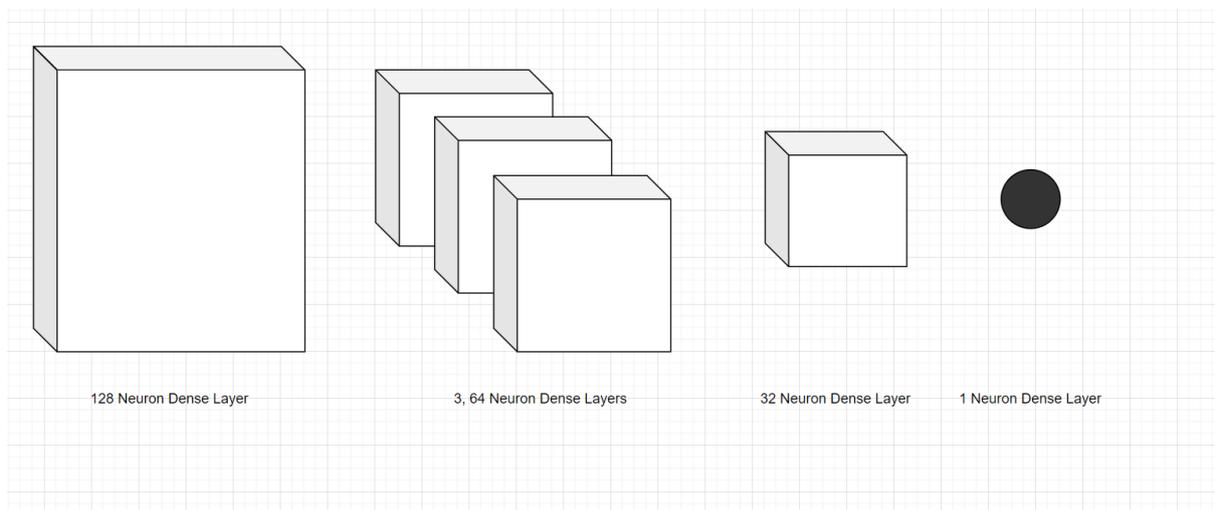

Figure 10: Architecture of the Deep Neural Network

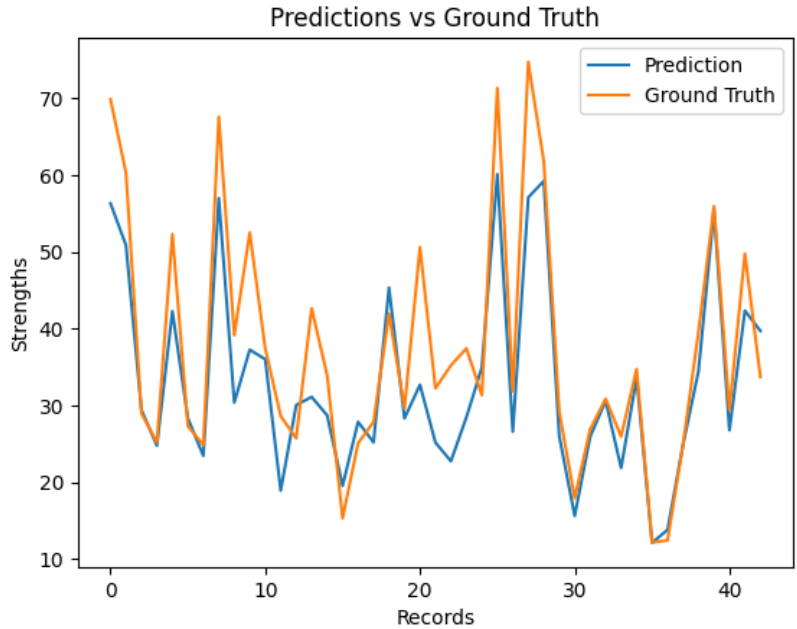

Figure 11: Predictions and corresponding ground truth values for DNN

### 5.4.1 Hyperparameter Tuning

For the DNN, *epoch* value is set at 100, with a *validation_split* of 0.05 This value can be increased if sufficient data is available. For model compilation, Adam optimizer, with a learning rate of 0.01 has been used.

## 6. Feature Engineering

Feature engineering can, therefore, greatly benefit us, especially if we have limited data, by engineering new features or modifying existing features for improving a model's predictive power. With more columns, some hidden patterns and relationships not otherwise captured by the original features will be brought forth, hence improving the model's accuracy and robustness. This process can also simplify the model by creating more informative features, thereby improving its generalization ability. Additionally, feature engineering allows for the incorporation of domain-specific knowledge, which can make the model more relevant and effective. Ultimately, it helps leverage existing data more effectively, augmenting the dataset and mitigating issues related to small sample sizes. Following steps have been take to perform feature engineering:

### 6.1 Outlier Removal

Sometimes, outliers have to be removed so that the statistical analysis or the machine learning model gives accurate and reliable results. In that respect, outliers might present misleading results, leading to misleading conclusions with turning-poor model performance. This might further bias the parameter estimate, raise the error rate, and decrease algorithm effectiveness in those assuming normality or homogeneity in data. On removing outliers, we get cleaner datasets, closer to what really happens, and hence improve model predictions and robustness of insights derived from the data. Proper outlier detection and handling also maintain dataset integrity and result validity. For outlier removal, z-score method (8) has been utilized

$$Z = \frac{X-\mu}{\sigma} \quad (8)$$

Here $\sigma$ is the standard deviation while $\mu$ represents the mean of distribution.

Any data point with z-score value greater than or equal to 3, or less than equal to 3 can be considered an outlier. Because of the negligible count of outlier data points, they have been deleted from the dataset.

## 6.2 Feature Addition

Second, since there is little data, one needs to exhaustively look into relationships between existing features and consider the creation of new features that might dramatically improve model performance. By doing so, from these relationships, we will be able to uncover latent patterns and interactions that are not directly observable but could make enormous improvements in predictive accuracy and robustness. To this end, the following relationships in the features have been recognized for further study and implementation into the model:

### 6.2.1 Superplastic, Water Relation

Superplasticizer and water follow an inverse relation. This can be observed from Figure 12 where the red line represents the best fit line.

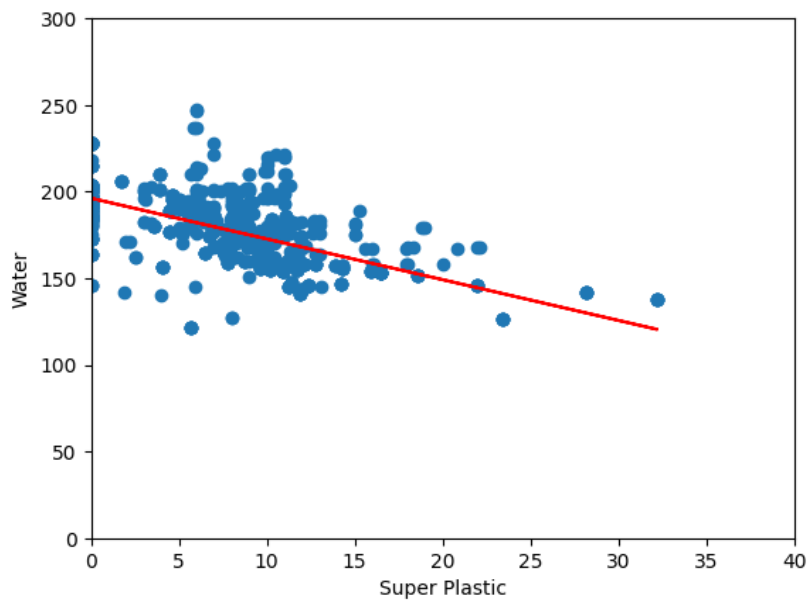

Figure 12: Water, Superplastic relation

Superplasticizers are kind of water-reducing agents applied in the process of concrete manufacturing. It reduces the viscosity of the mix, hence making it easier and more convenient to shape. Doing so shall substantially reduce the water-to-cement ratio required, thereby resulting in stronger and more resilient concrete. With less water, the concrete becomes denser and less porous; it is now better at withstanding natural challenges such as freeze-thaw cycles, unwanted chemical reactions, and abrasion. Consequently, improved compressive strength and higher durability mean better concrete performance.

Furthermore, superplasticizers ensure desired concrete properties with advanced slump and enhanced flowability. The required flow properties are found, for example, in complex formworks and densely reinforced structures. Improvement of workability by adding superplasticizers does not involve losses concerning mechanical properties, which makes this component very important for modern concrete production.

### 6.2.2 Strength, Water Cement Ratio Relation

An interesting observation made is the relation of strength and water: cement ratio. Figure 12 highlights the relation discovered from the dataset:

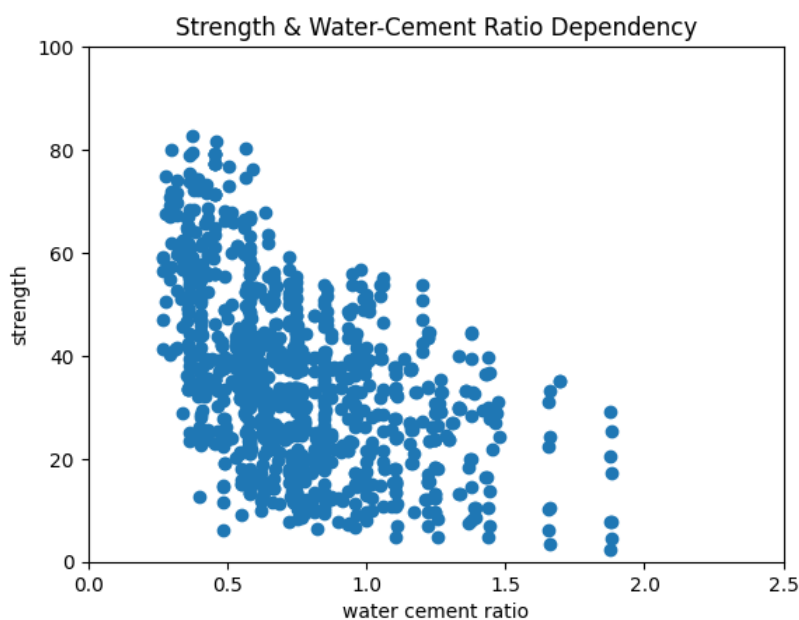

Figure 13: Relation of strength and water: cement ratio

From Figure 13, a hyperbolic relation of the 2 features can be observed. Figure 14, illustrates hyperbolic shape

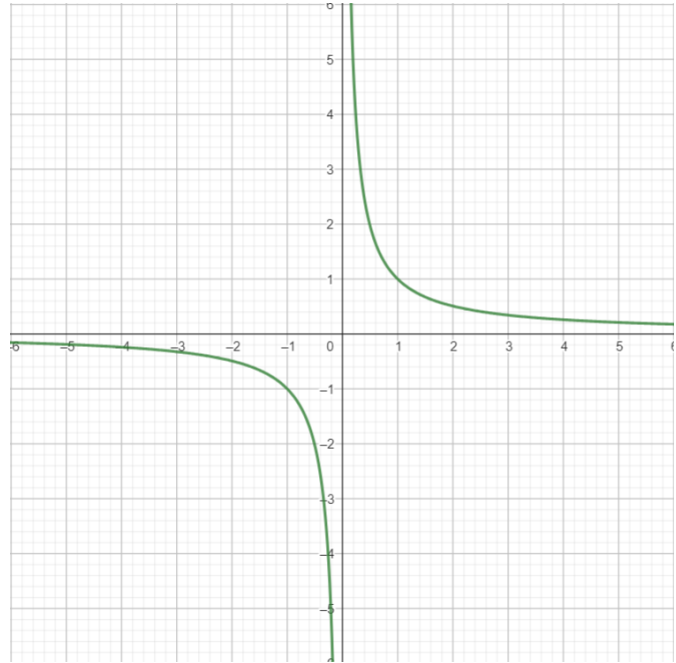

Figure 14: Hyperbolic curve ($y = \frac{1}{x}$)

This hyperbolic relationship is no coincidence; it's a well-established concept, as confirmed by research conducted at the University of Illinois [14]. The study shows that a low water-to-cement ratio results in high strength but reduces workability. Conversely, a high water-to-cement ratio improves workability but at the cost of reduced strength.

This result led to introducing a new feature in the dataset: water: cement

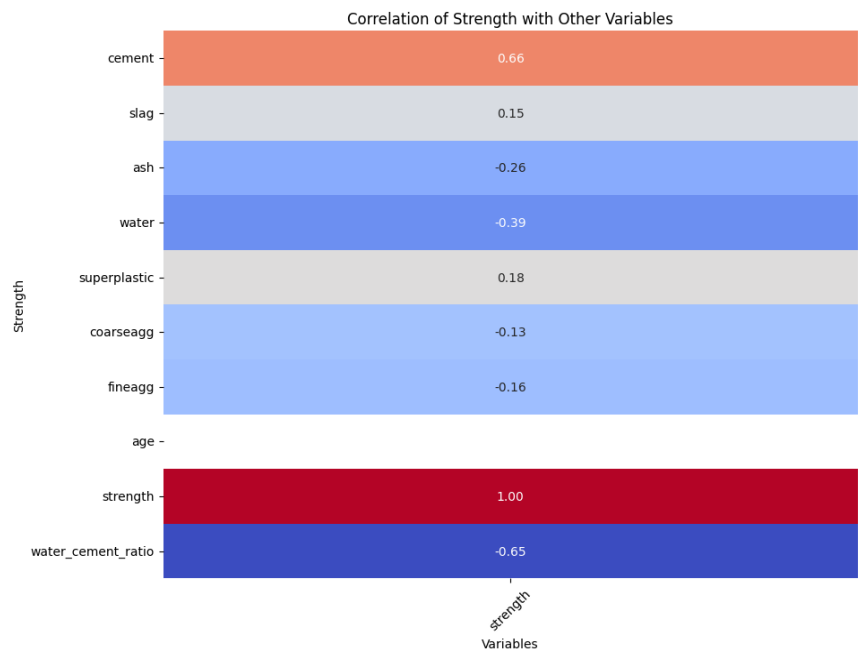

Figure 15: Correlation of input features with strength

From the Figure 15 it can be seen that the new feature water: cement ratio has the highest negative correlation. This makes ensures us that adding the new feature was indeed beneficial for predicting the strength.

### 6.2.3 Cement, Strength Relation

Figure 14 illustrates the relation of cement and strength of concrete.

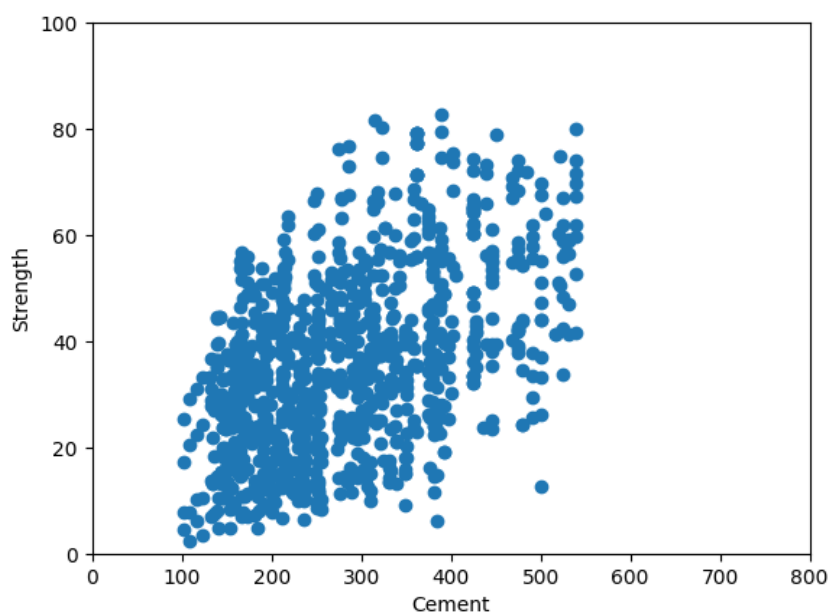

Figure 16: Cement, strength relation

From Figure 16, a positive correlation can be observed within the 2 features. Unfortunately, the correlation coefficient for this relation is less, which makes us incapable of utilizing it for enhancing the model performance.

## 6.3 Standardization

Standardizing features is crucial for improving model performance because it ensures that all features contribute equally, rather than allowing any single feature to dominate due to differences in scale. By adjusting the features so their mean value becomes zero with a standard deviation of one, standardization aligns the scales, making it easier for the model to learn efficiently. This is especially important for algorithms like gradient descent, which operate more effectively and converge faster when features are on a similar scale. Ultimately, standardization leads to more reliable, robust, and interpretable models.

For standardizing the features, z-score method mentioned in the outlier removal topic has been used.

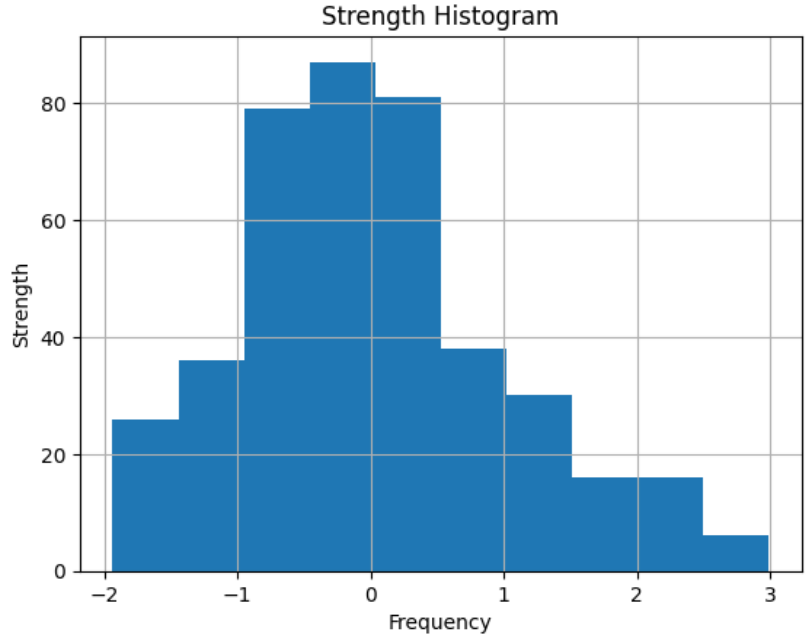

Figure 17: Distribution of strength feature after standardization

After performing the feature engineering, all the models were retrained on the new dataset. It was noted that on an average, there was a 20.725% decline in the RMSE loss values which proves that the efforts bore fruits.

## 7. Physics Informed Neural Network (PINN)

PINNs belong to the category of neural networks that utilizes both physics, and the power of neural networks to make better predictions. Unlike traditional neural networks, which are based on purely data-driven techniques, in PINNs, the differential equations and required physical constraints are directly included in the loss function of the model. This will enable the model to be trained in such a way that known physical laws, such as mass or energy conservation, are guaranteed to be satisfied, while still being able to learn from data. It becomes especially appropriate for the solution of complicated scientific and engineering problems in which there is very limited data but physical principles are quite well understood. In that respect, PINNs can give better and more reliable predictions by using empirical data and fundamental physics. This approach thus turns out to be very powerful in many areas of fluid dynamics, structural analysis, and materials science.

### 7.1 Abrams Law

For creating PINNs, its necessary to learn the physical formulas defining the strength of resultant concrete. The formula would be required to alter the loss function of DNN model to transform it to PINN. One of the popular equations defining strength is [15] Abrams Law. Abrams law states:

$$concrete\_strength = \frac{A}{B^{\frac{water}{cement}}} \qquad (9)$$

In the above formula, A, and B are constants. Water and Cement must be replaced with respective quantities of those materials. In order to get any useful predictions from the Abrams Law, the values of A, and B must be figured out. For finding these values, Heuristic Optimization method must be applied, so that the strength values from the Abrams Law could be as close to the ground truth values as possible.

## 7.2 Particle Swarm Optimization for Abrams Law

If we take any random value of A, and B, say 10 and 4, then Figure 18 illustrates the variation between both the quantities graphically:

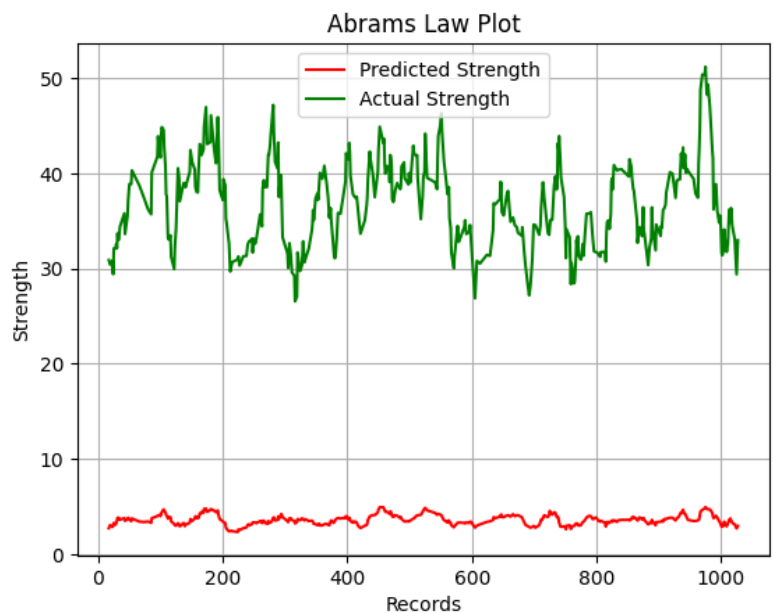

Figure 18: Strength from Abrams Law, and the ground truth

Kindly note that moving average has been applied multiple times in Figure 17 to smoothen it, and make the variations with ground truth clearly visible. In order to find the values of A, and B that can get the 2 curves of Figure 17 overlap much better, we shall be applying Particle Swarm Optimization algorithm. Particle Swarm Optimization (PSO) is a heuristic optimization technique which involves the concept of creating virtual particles that tend to move to the location/ state that return the best cost function value. The movement of these particles depends on two factors: best position they discovered and the best position discovered by the entire swarm. This approach helps them gradually converge toward optimal solutions over successive iterations. PSO is widely used in tackling complex optimization problems in various fields such as engineering design, machine learning hyper-parameter tuning, and financial modelling. Its popularity stems from its simplicity, flexibility, and ability to efficiently navigate non-linear, multi-dimensional search spaces. After applying Particle Swarm optimization, the values of coefficient A, and B comes out to be 81.58897468 and 2.7157893

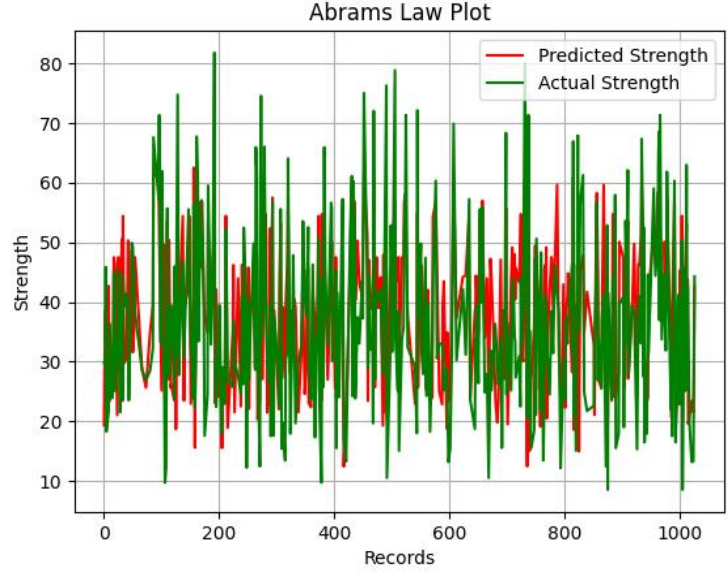

Figure 19: Strength from Abrams Law, and ground truth

The RMSE loss value of the PSO model in Figure 18 was 35.86724 while the loss value in Figure 19 was reduced to 10.75657 which is 3.33 times less than usual! PSO applied here, has a particle count of 40, and the cost function computes the RMSE loss value.

### 7.3 Altering DNN Loss Function for PINN

The loss function used in the ordinary DNNs contain only *Data Loss* term. In our case, RMSE loss term is present there. In order to convert it to a PINN, an additional term must be added. This loss term is called *Physics Loss*. This term must calculate the loss value utilizing a physics formula. This method makes the model utilize the existing knowledge of physics and work better even with reduced dataset. The physics loss term developed for PINN is denoted by the equation (10)

$$\text{Physics Loss} = \sqrt{\sum_{i=0}^{i=N} \frac{(y_i - \frac{A}{\frac{water}{B cement}})^2}{N}} \qquad (10)$$

The term above is the Root Mean Squared of the difference of strength predicted by Abrams law, and that predicted by the DNN. Hence the final loss function comes out to be equation (11)

$$\text{Loss} = \sqrt{\frac{\sum_{i=0}^{N}(y_i - \hat{y}_i)^2}{N}} + \sqrt{\sum_{i=0}^{i=N} \frac{(y_i - \frac{A}{\frac{water}{B cement}})^2}{N}} \qquad (11)$$

The first term is the data loss function while the 2nd term is the physics loss term. This is how the DNN has been converted to a PINN.

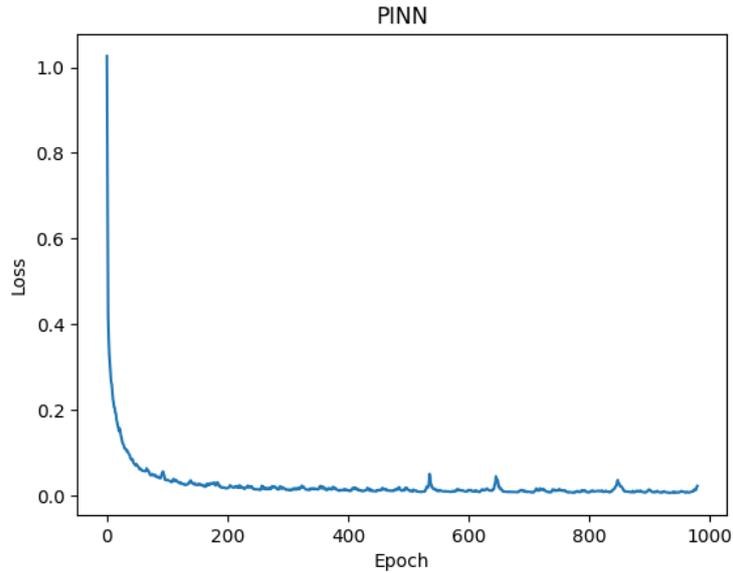

Figure 20: Loss values corresponding to the epoch count

The RMSE loss value using PINN came out to be 3.908 This is the least value encountered till yet, and is approximately 26% less than the conventional DNN. Thus, utilizing a combination of Abrams' Law and DNN helped to significantly reduce the loss value. Abrams' Law alone wouldn't have fulfilled the objective of obtaining the most accurate predictions, nor would DNN alone.

## 7.4 Performance With Reduced Dataset

One of the major reasons of utilizing PINNs is its unique ability to perform equally well with reduced dataset. Availability of dataset has always been an issue for these data-hungry models. After training the PINN model, it became evident that even with up to 40% less data, PINN worked better than the conventional DNN. After 40% data reduction, its performance gradually starts to decrease.

## 8. Heuristic Optimization for Cost Minimization

Predicting strength won't be much useful for any organization. The difficult part is to predict the quantity of raw materials for a given threshold strength of concrete, and minimizing the cost of procurement. For the same, we shall be utilizing PSO, and the strength prediction model developed before. This shall enable us to minimize the cost while keeping the strength maximum.

## 8.1 Particle Swarm Optimization Specifications

For implementing the PSO algorithm for predicting the quantity of raw materials, the following details must be known before implementing it:

### 8.1.1 Distribution of Features

In order to predict the quantity of each raw material required, it is necessary to set an initial bound of these features which would reduce the search space, and speed up the process. For setting the initial bounds, the distribution of features has been studied, and the most commonly occurring range is selected as bounds. For now, we shall be considering the threshold value of the strength of concrete that the client might require to be 30 This means that the distribution indicated in Figure 21 is of only those records where the output strength was greater than 30

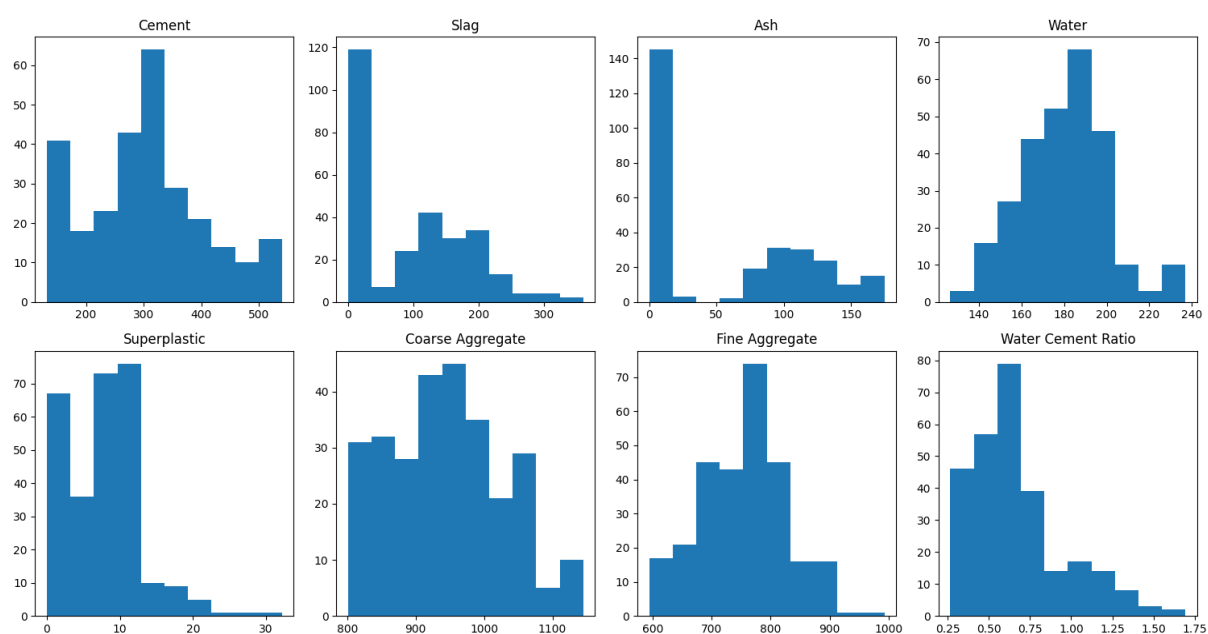

Figure 21: Distribution of feature, where concrete strength ≥ 30

### 8.1.2 Setting Cost Function

For the PSO to work, a cost function is required that could somehow quantitatively score the solution and help the PSO algorithm to keep enhancing the score to reach the best, and the most optimized solution. For this reason, the cost function developed takes into account the following factors:

- For encouraging the algorithm to suggest combinations for concrete where threshold strength is 30, a reward must be given to it, if the solution that it suggests has a resultant strength of at least 30
- In order to minimize the cost, the algorithm must be rewarded if the combination it suggests has a cost less than the average cost of records in the dataset where strength <30
- If the algorithm fails to fulfill any of these, a punishment must be given. (negative reward)

### 8.1.3 Setting Other Necessary Parameters

For PSO, the count of particle set was 40 For systems with greater computational power, this count can be increased, which might provide better results. Additionally, the count of variables to be optimized is set at 8 (7 input features + water: cement ratio feature)

After executing the PSO algorithm, the quantity of raw materials suggested generally had a strength in the range of 50-60, and cost 5-10% less than the average. This phenomenal performance could be of great advantage to the consulting industries.

9. Results

Table 3: Comparison of performance before and after feature engineering

| Loss Before | Loss After | Performance Improvement | Model |
|---|---|---|---|
| 7.331 | 6.91 | 5% | Linear Regression |
| 9.62 | 5.7 | 40.7% | Random Forest |
| 8.377 | 7.31 | 12.7% | Gradient Boosting |
| 7.03 | 5.303 | 24.5% | DNN |

Table 4 compares the performance of the model before and after feature engineering. From that, it can be concluded that feature engineering helped with reducing the loss values of the model by an average 20.725% Random Forest showed the highest improvement of 40.7% DNN still ranks the best among rest of the models because of its least RMSE loss value.

Table 4: Performance of PINN with reduced data

| Data Reduction | PINN Loss (RMSE) | Ordinary DNN Loss |
|---|---|---|
| 10% | 3.908 | |
| 20% | 5.177 | |
| 30% | 5.381 | |
| 40% | 5.66 | 7.03 |
| 50% | 7.423 | |
| 60% | 7.04 | |
| 70% | 7.53 | |

Furthermore, the integration of Physics-Informed Neural Networks (PINNs) substantially boosted the performance of the Deep Neural Network, even with up to 40% less data. This approach addressed one of the most pressing issues faced by consulting agencies: limited data availability. By incorporating PINNs, not only was the quantity of data required reduced, but the loss value also decreased by 26.3%, significantly improving the model's efficiency and reliability.

Additionally, the Particle Swarm Optimization (PSO) algorithm exceeded expectations by reducing the procurement cost of raw materials by 5-10%. Notably, it also enhanced the strength of the output, with values surpassing 50, well above the threshold strength value of 30. This remarkable advancement can greatly assist industries in lowering production costs while maintaining high product quality.

## 10. Conclusion

The PINN model introduced in this paper marks a major advancement in machine learning by seamlessly integrating traditional data-driven methods with the extensive body of knowledge that humanity has developed over centuries. This approach enhances the predictions and also ensures that the model aligns with well-established physical laws, creating a powerful tool that leverages both empirical data and fundamental scientific principles. While conventional neural networks rely solely on data to make predictions, they often struggle when data is scarce or incomplete, leading to inaccuracies and unreliable results.

This is where the brilliance of the PINN model shines through. By embedding the fundamental principles of physics directly into the neural network, the model gains a deeper understanding of the system it is trying to predict. This integration of physical laws allows the PINN to make more accurate predictions, even when faced with limited or noisy data. In essence, the model compensates for the lack of data by leveraging the well-established knowledge of physics, ensuring that the predictions are not just data-driven but also grounded in the reality of how the world works.

The importance of this approach cannot be overstated. In many real-world scenarios, collecting vast amounts of data is either impractical or impossible. Traditional models falter in such situations, but the PINN model's ability to incorporate physical laws offers a robust solution. By doing so, it not only enhances the accuracy of predictions but also broadens the applicability of neural networks to areas where data limitations have previously been a significant barrier. This marriage of physics and machine learning represents a promising path forward in creating models that are both intelligent and insightful, capable of overcoming the very limitations that have long challenged data-driven methods.

## 11. Limitations and Future Direction

The development of the current PINN model represents a significant advancement in predicting concrete strength, primarily using Abrams' Law as a foundational principle. While this approach has proven effective, additional factors, such as temperature during mixing and placing, and the specific type of cement used, also play crucial roles. These factors were not included in the current model due to the lack of available data. This limitation provides a clear and exciting direction for future research.

The research team is enthusiastic about the next phase, which involves creating more refined models that incorporate these additional variables and laws. This will require comprehensive data collection and model refinement to ensure that all relevant factors are considered, thus improving prediction accuracy and reliability. The adaptation of ordinary DNNs into PINNs has been complex, requiring expertise in tensor operations and loss function customization, but the successful implementation demonstrates its feasibility and promise.

Looking ahead, the goal is to integrate other critical features influencing concrete strength that were not utilized in the current PINN version. By expanding the model's scope to include a broader range of data, the researchers aim to develop a more robust and comprehensive tool for predicting concrete strength. This endeavour highlights their commitment to advancing the field and underscores the potential for PINNs to provide valuable insights and solutions in concrete manufacturing and construction.

## 12. Ethical Statement

This research was conducted at KPMG India, with the support of the Industry 4.0 team. All necessary permissions for conducting the research and publishing the results independently have been obtained from KPMG India. The dataset utilized in this study was sourced from Kaggle and is licensed under the CC0 (Creative Commons Zero) license, allowing for unrestricted use and distribution. All rights to use and publish the data were granted in accordance with the dataset's license.

The authors declare that the research adhered to ethical standards, ensuring transparency, integrity, and accountability throughout the study. The authors did not receive any external funding, and there are no conflicts of interest to disclose. This research was conducted with the intent to contribute positively to the academic community and industry practices, without causing harm to any individuals or groups.